\documentclass[runningheads]{llncs}
\usepackage{graphicx}
\usepackage{hyperref}

\usepackage{subcaption}

\usepackage[color=pink]{todonotes}
\usepackage{xspace}
\usepackage[capitalize,noabbrev]{cleveref}
\usepackage{booktabs}

\newcommand{\sig}{\textsc{Sigmorphon 2016}\xspace}
\newcommand{\japan}{\textsc{Japanese Bigger Analogy Test Set}\xspace}

\begin{document}
\title{On the Transferability of Neural Models of Morphological Analogies\thanks{This research was partially supported by TAILOR, a project funded by EU Horizon 2020 research and innovation programme under GA No 952215, and the Inria Project Lab ``Hybrid Approaches for Interpretable AI'' (HyAIAI).}}
%
%
\author{Safa Alsaidi\inst{1}\orcidID{0000-0002-4132-1068} \and
Amandine Decker\inst{1}\orcidID{0000-0001-6773-9983} \and
Puthineath Lay\inst{1}\orcidID{0000-0002-0850-8656} \and
Esteban Marquer\inst{1}\orcidID{0000-0003-2315-7732} \and
Pierre-Alexandre Murena\inst{2}\orcidID{0000-0003-4586-9511} \and
Miguel Couceiro\inst{1}\orcidID{0000-0003-2316-7623}}
\authorrunning{S. Alsaidi et al.}
%
\institute{Université de Lorraine, CNRS, LORIA, F-54000, France\\
\email{\{esteban.marquer,miguel.couceiro\}@loria.fr} \and
HIIT, Aalto University, Helsinki, Finland\\
\email{pierre-alexandre.murena@aalto.fi}}
\maketitle              

\begin{abstract}
Analogical proportions are statements expressed in the form ``A is to B as C is to D'' and are used for several reasoning and classification tasks in artificial intelligence and natural language processing (NLP).
In this paper, we focus on morphological tasks and  we propose a deep learning approach to detect morphological analogies. We present an empirical study to see how our framework transfers across languages, and that highlights interesting similarities and differences between these languages.
In view of these results, we also discuss the possibility of building a multilingual morphological model.    


\end{abstract}

\section{Introduction}
\par An analogy, or analogical proportion, is a relation between two word pairs or four elements $A$, $B$, $C$, and $D$, meaning ``$A$ is to $B$ as $C$ is to $D$'', often written as $A:B::C:D$. An analogical proportion becomes an equation if one of its four objects is unknown \cite{Miclet_2008}. 
\par Analogies have been extensively studied in Natural Language Processing, which resulted in different formalizations with noteworthy applications in various domains such as derivational morphology \cite{ijcai2020-0256}. Analogies on words can refer exclusively to their morphology as in the following example: ``\textit{apple} is to \textit{tree} as \textit{apples} is to \textit{trees}''. It is based on morphological variations of two words: ``apple'' and ``tree''. The question of the correctness of an analogy $A:B::C:D$ is a difficult task; however, it has been tackled both formally and empirically \cite{ijcai2020-0256,lim2019,2003:lepage}. Recent empirical works propose data-oriented strategies to learn the correctness of analogies from past observations. These strategies are based on machine learning approaches.

Analogical classification and inferences are two main problems worth addressing when dealing with the axiomatic settings of analogies. Multiple attempts have been conducted in terms of formulating and manipulating analogies \cite{analogy-formal:2004:lepage,2003:lepage}. Most algorithmic approaches to solving morphological analogies rely on the formal characterization of proportional analogies (see \cref{sec:exp}).

However, other approaches to detecting analogies have been used as well. Some of which include Fam and Lepage (2018)~\cite{tools-analogy:2018:fam-lepage} and the Alea algorithm by Langlais \textit{et al.} (2009)~\cite{analogy-alea:2009:langlais}.
Fam and Lepage's approach relies on feature vectors to detect analogies within a list of words and create analogical grids. As a result, analogies between more than 4 words can be generated. The axioms of Lepage \cite{analogy-formal:2004:lepage} have been also reformulated by Yvon \cite{finite-state-trancducers:2003:yvon} to give a closed form solution, which is computed by Alea in a Monte-Carlo setting, where character strings $A$, $B$ and $C$ are randomly sliced and merged to obtain potential solutions to $A:B::C:x$. 

An approach by Murena \textit{et al.} (2020)~\cite{ijcai2020-0256} outperformed both  Alea and Fam and Lepage's approaches. In this approach, analogical equations $A:B::C:x$ are solved by finding the $x$ that minimizes the Kolmogorov complexitity, which is evaluated by using a simple description language for character strings and an associated binary code.

Another approach by Lim \textit{et al.} \cite{lim2019} propose to learn an analogy operator directly from analogies. They used neural networks to learn semantic analogies, and proposed different models for analogical classification and inferences (regression). They also used pretrained GloVe embeddings \cite{glove:pennington}, which achieved competitive results on analogy classification and completion. Moreover, they relied on the properties of formal analogies to increase the amount of training data as introduced in \cref{sec:data-augment}. 

In this paper, we adapt the approach developed by Lim \textit{et al.} \cite{lim2019} for semantic word analogies and apply it on morphological analogies. For this approach, we had to develop and train various morphological word embedding models. Furthermore, we explore the potential of transferring our neural analogy model across 11 languages, which is the core of this paper. Our method achieves competitive results on each language and promising ones in the transferability settings. This allowed us to explore building a single multilingual model that could work with several languages.

This paper is organized as follows. In this introduction, we recall some of the related works particularly the one by Lim \textit{et al.} \cite{lim2019} from which our framework was inspired. In \cref{sec:approach}, we explain how we adapt this approach to morphological analogy and the models we developed and trained. In \cref{sec:exp}, we introduce our datasets and describe how the models trained on each language perform when transferred to the other languages. We further investigate different transferability settings in \cref{sec:lan_model}. We work with 4 language models that reveal interesting results.

\section{Proposed Approach} \label{sec:approach}
As aforementioned, we adapt the approach by Lim \textit{et al.} \cite{lim2019} in terms of using the same architecture for our classification model. However, we have developed our own custom embedding models, which is trained along with the classifier.

\begin{figure}[tbp]
\centerline{\includegraphics[width=0.85\textwidth]{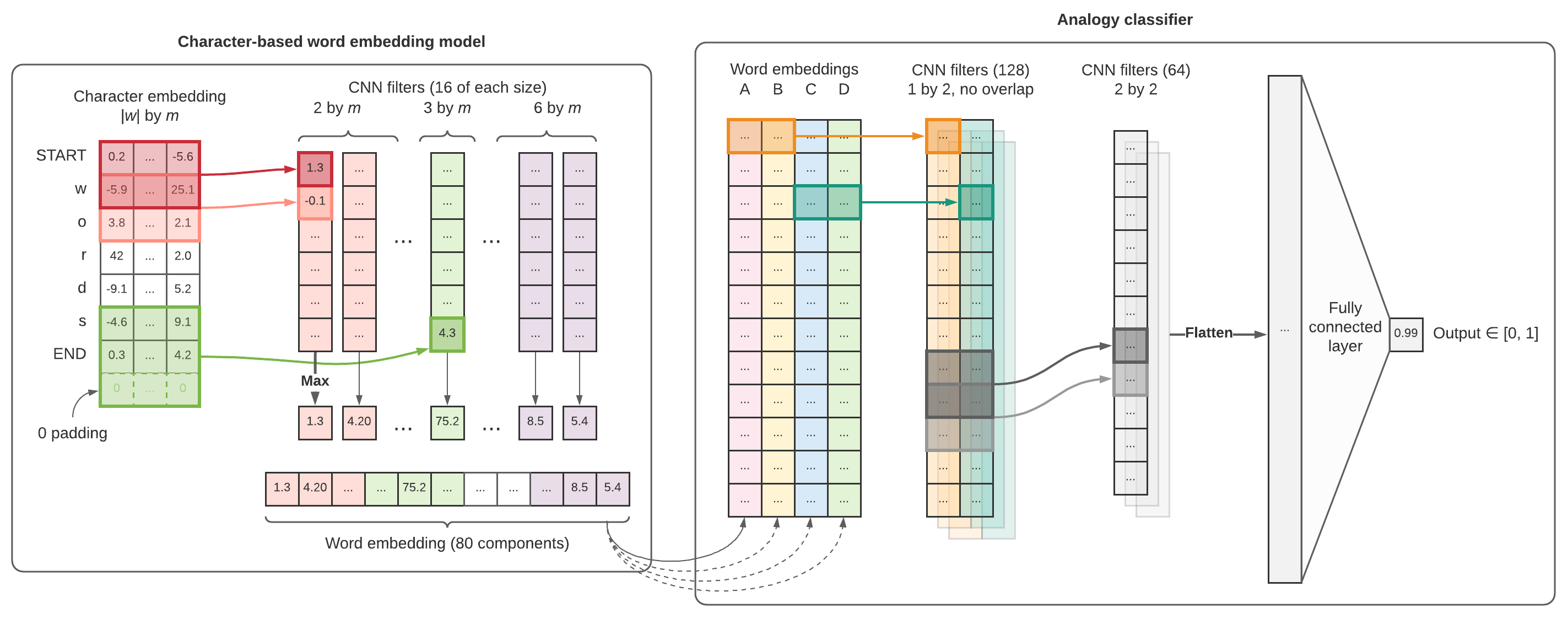}}
\caption{The character-level word embedding model and the  CNN classifier.}
\label{fig:char-embedding}
\end{figure}

\subsection{Classification Model}
The architecture of the classification model is a Convolutional Neural Network (CNN), which takes as input the embeddings of size $n$ of four elements $A$, $B$, $C$ and $D$ stacked into a $n$ by $4$ matrix. The CNN has 3 layers as follows (see \cref{fig:char-embedding}).  A first convolutional layer with filter of $1$ by $2$ is applied on the embeddings. This layer respects the boundaries between the two pairs, where it analyses each pair and extracts the differences and similarities between $A:B$ and $C:D$. We use 128 of such filters with a Regularized Linear Unit (ReLU) as activation function.
 
  A second convolutional layer with 64 filters of $2$ by $2$ is applied on the resulting $128$ by $n$ by $2$ tensor, after which the result is flattened into a $64 \times (n-1)$ unidimensional vector. ReLU is also used as an activation function. This layer should compare the results of the analysis of both pairs: if $A$ and $B$ are different in the same way as $C$ and $D$, then $A:B::C:D$ is a valid analogy.
  In the third layer, the results are flattened and used as input of a dense layer. We use a sigmoid activation to get a result between $0$ and $1$ as we work with a binary classifier.

\subsection{Embedding Model}\label{sec:emb}
In  \cite{lim2019}, Lim \textit{et al.} achieved good results on semantic analogies by using pretrained GloVe embeddings that are supposed to encode the semantics of words. But when we tried to apply it to morphological analogies, we got poor performance where after investigation it turned out that the model could not learn but rather produced tensors full of zeros. Therefore, since we work with morphological analogies and with languages with no embedding models already available, we had to develop and train our own models. We developed a character-level word embedding model able to capture the morphological aspect of words. To our knowledge no such embedding model is available for the languages we manipulate, so we decided to train an embedding model together with our classifier. We use a very simple model architecture based on Convolutional Neural Network (CNN) as described in \cite{morpho-thesis:2020:vania}. This architecture is designed to embed individual words for morphological tasks. In our case it is composed of two layers as follows.

First, we have a character embedding layer to encode each character of the word with a vector of size $m$. Characters never encountered during the training phase are embedded with vectors full of 0 by default. At the beginning and at the end of the word, we add special vectors to signify the boundaries of the word. For a word of  $|w|$ characters, we obtain a $(|w|+2)$ by $m$ matrix.

Then we use a convolutional layer with 16 filters of each of the sizes $2$ by $m$, $3$ by $m$, \ldots,  $6$ by $m$, for a total of $16\times 5=80$ filters.
The filters are used on the embeddings so that they overlap along the character dimension. We expect those filters to capture morphemes of up to 6 characters.

A max pooling layer is then  applied to the output where we keep only the greatest number produced by each of the $80$ filters so that only the most important patterns appear in the final embedding. We finally concatenate the results to produce an embedding of size $80$.

\section {Experiments}\label{sec:exp}
We worked with 11 languages in our experiments:  10  from \sig \cite{cotterell-sigmorphon2016} and the other  from \japan \cite{KarpinskaLiEtAl_2018}.
\subsection{Datasets and Augmentation}\label{sec:data-augment}
The \sig \cite{cotterell-sigmorphon2016} datasets contains training, development and test data. Data is available for 10 languages: Spanish, German, Finnish, Russian, Turkish, Georgian, Navajo, Arabic (Romanized), Hungarian and Maltese (see \cref{tab:sigmorphon} in Appendix). It is separated in 3 subtasks: inflection, reinflection and unlabeled reinflection. In our experiments, we focused on the data from the inflection task, which is made up of triples $\langle A, F, B\rangle$ of a source lemma $A$ (ex: ``cat''), a set of features $F$ (ex: pos=N,num=PL) and the corresponding inflected form $B$ (ex: ``cats''). On the other hand, Japanese dataset contains several files, each of them containing pairs of linguistically related words. We were interested in inflectional and derivational morphology relations for which the dataset contains respectively 515 and 502 pairs of words. For each two pairs with the same relation, we produced an analogy. However, Japanese produced poor results when solving analogies which may be related to the small size of the dataset.

 Deep learning approaches require large amounts of data. Therefore we took advantage of some properties of analogies to produce more data based on our datasets, this process is called \textit{data augmentation}. Given a valid analogy $A:B::C:D$, we can generate 7 more valid analogies, namely, $A:C::B:D$, $D:B::C:A$, $C:A::D:B$, $C:D::A:B$, $B:A::D:C$, $D:C::B:A$, $B:D::A:C$, and 3 invalid analogies, namely, $B:A::C:D$, $C:B::A:D$, and $A:A::C:D$.\footnote{Lim \textit{et al.} \cite{lim2019} generate the 3 invalid analogies for each of the 8 valid analogies, resulting in 24 invalid analogies. We experimented with 24 invalid analogies but the performance of the model was worse than when using only 3 invalid analogies, despite the increase in the amount of training data. The decrease in performance is likely due to the stronger imbalance between invalid and valid examples.}
 
 

\subsection{Transferability Results}
 In this section, we will explain the results we obtained once we applied the different models trained on each language to the other languages of the dataset. The objective is to explore the generalization capabilities of the CNN model, and to test its dependence on the training language. To evaluate their generalization capacity, we ran the evaluation using each model on all the languages. In the next subsections we introduce our results for full and partial transfer. In full transfer, we describe the results when transferring both the embedding and classification models. Then in partial transfer, we transfer only the classification model (the data is embedded with the ``right'' embedding model). We test them on 3 settings: base, negative, and positive analogies. Positive and negative analogies stand for the valid and invalid forms of analogies, which are presented in the previous section. Base is the initial analogy form \textit{$A:B::C:D$}, we do not use the data augmentation for this evaluation.
 

\subsubsection{Full Transfer} \label{sec:full_transfer} 
The results for full transfer (\textit{i.e.,} using both the embeddings and the classifier of a language on another language) on positive and negative data are presented in \cref{fig:cmap-full} in Appendix. The results for positive data are most of the time above 90\% except for Arabic and Navajo words, compared to the results of negative data, which were more heterogeneous. This is probably due to differences in alphabet between the source and target language that cause a large portion of the target alphabet to be unrecognized. As a result, characters unknown to the model are embedded as zeros and tend to be ignored by the model. The very low performance on negative classification can be explained in that the classifier answers ``valid analogy'' by default when encountering unrecognized characters. 

Overall, most languages transfer reasonably well, with an accuracy between 50\% to 80\%.
However, models have an accuracy close to 0\% when transferred across languages that don't share most of the alphabets. Interestingly, models trained on Turkish and Hungarian perform slightly better when compared to those trained on other languages. Further analysis and experiments are introduced in \cref{sec:lan_model}.

\subsubsection{Partial Transfer} \label{sec:partial_transfer}
To solve the issues we had with character dictionaries for full transfer, we tried to transfer only the classifier. The results for partial transfer (\textit{i.e.,} using the classifier model of a language on another language using the target language's embedding model) are presented in \cref{fig:cmap-part} in Appendix.

The results are similar to those of full transfer, where the model transfers well for positive data but results remain heterogeneous for negative data. As we used the embedding model corresponding to the language, we have less combinations producing an accuracy of 0\% for negative data. To elaborate, the performance is above 10\% for all but three cases (from Georgian and Spanish to Japanese and from Spanish to Arabic), and above 25\% in most cases, which is a clear improvement from full transfer.

The main difference between both experiments is that for partial transfer, the alphabet gap is no longer an issue which results in a significant improvement in the performance. But since the embedding model was not trained together with the classifier, this may result in a mismatch in the representation. We tried to address this potential issue with our experiments with multilingual models.

\subsection{Discussion}
The quality of the transfer is not symmetric, which means that a model that performs well when transferred from a language~$A$ to a language~$B$ may perform badly when trained on language~$B$ and transferred to language~$A$. The Hungarian model transfers really well to other languages while most of the other languages are not that efficient with Hungarian. This could be explained due to the fact that Hungarian has ``particularly rich morphology'' \cite{hungarian} using inflection, derivation and compounding.

The Arabic model is efficient for full transfer but not really for partial transfer. It could be possible that the Arabic embedding model encodes the sub-words differently from the other models, as Arabic words are formed by roots and word patterns. This would mean that the embedding and classification model would be strongly related. As a result, it performs poorly once applied to other languages in case of partial transfer. 

For Japanese, the accuracy is not always of 0\% for negative data when the models transfer to Japanese. None of the Japanese characters are present in the dictionaries of the other models so we could expect all the analogies to look like $\varepsilon:\varepsilon::\varepsilon:\varepsilon$ and thus all the said invalid ones to be classified as valid. The reason why we do not have 0\% in those cases is still unknown.

Eventually Hungarian seem to be very efficient models in terms of transfer learning. In average, Hungarian is one of the languages closest to all the others in the family hierarchy, which may explain why it transfers well. As a result, we decided to run more experiments to further evaluate and improve our models in terms of transferability. Further details are introduced in the next chapter.

\section{Toward a Multilingual Model} \label{sec:lan_model}
Following our observations, we decided to work on building a multilingual model. Our idea was that using several languages during the training phase would increase the generalization capacity of our model. We thus trained different models with different languages as training data in order to see which setting produces the best results in terms of transferability. We then evaluated the transferability of the models across the 11 languages to determine if we obtain better performance than with only one language during the training phase.

\subsection{Models}
We explored two different settings regarding the languages we used: on the one hand we trained models with two languages as training data and on the other hand we trained models with all our eleven languages or 10 without Japanese.

\subsubsection{Models with two languages} \label{sec:bilingual}
For our bilingual models, we worked with two pairs of languages: Hungarian-Finnish and Hungarian-Turkish. We chose Hungarian and Finnish because they are close to all the other languages in the family hierarchy. As for Hungarian and Turkish, they are the languages that produced the best results in terms of transferability.

\subsubsection{Models with all languages}\label{sec:multi}
There are two types of multilingual models. In one case, we trained a \emph{single embedding model} for all the languages. We thus use data of all the languages. Regarding the training set data, it is shuffled so that the model is not trained on one language after the other. In the other case, we train one embedding model per language (called \emph{multi-embedding model}); therefore, there will be 10 or 11 embedding models for 10 or 11 languages model. For this model, the training sets of data are concatenated one after another. 


\subsection{Results}

\subsubsection{Models with two languages}{}
As we could expect, for full transfer, the results (see \cref{fig:cmap-biling} in Appendix) turned out to be very good when transferred along the same languages (\textit{i.e.,} Hungarian and Finnish when applied to Hungarian or Finnish produces good results and same for the other model). The results were very bad for Russian and Georgian (\textit{i.e.,} the accuracy is around 2\%) and for Japanese (\textit{i.e.,} the accuracy is 25\%) when the Hungarian and Finnish model is applied. Navajo and Arabic is slightly worse in full transfer compared to other languages for both models.

For partial transfer, applying the Hungarian and Finnish model to the Hungarian or Finnish gave slightly worse results as each of the individually trained embedding model are not being perfectly adapted to the classifier. However, the results for the Hungarian and Turkish were very good when applied to Hungarian or Turkish. For positive (valid) examples, the results of the Hungarian and Finnish model was very good when applied to the rest of the languages. It was slightly lower for negative examples (\textit{i.e.,} the accuracy is still above 30\% for invalid), except for Japanese. In comparison, the results for the Hungarian and Turkish model were very strange when applied to the rest of the languages. Lower performance can be observed on negative examples for Georgian and much lower performance on Japanese between base and positive examples.





\subsubsection{Models with all languages}
In the experiment, we train on both 10 languages and 11 languages (with and without Japanese). So, we train 2 models of single embedding model and others 2 of multi-embedding model (see \cref{sec:multi}). The accuracy results (see \cref{fig:cmap-multiling} in Appendix) of the four models are comparable, except for Maltese, where the performance is lower than those of other languages. Both models with Japanese obtain better results, especially on invalid examples. The single embedding model with Japanese provides the best result, especially the result with 91\% of Japanese invalid accuracy.


\subsection{Discussion}
Overall, the bilingual models achieve better results than the multilingual ones. However, the bilingual models produce low results for a few languages while the multilingual models are more constant. Moreover, we used less than 5000 analogies per language to train the multilingual models, while we use 25000 analogies per language for the bilingual models. We could expect the performance of our multilingual model to increase with more training data.

During these experiments, we also tested another training setting. Instead of using 8 valid analogies for 3 invalid ones, we applied the properties on the 3 invalid forms. In the end we had 8 valid analogies for 24 invalid ones. However this setting produced worse results than the one with 3 invalid analogies. This could be related to the stronger imbalance between the positive and negative data. To solve this issue, it would be interesting to train models on balanced datasets: we would work with the 8 valid forms and 8 invalid forms randomly selected among the 24.

Eventually we tried to perform an analysis of the similarities between our 11 languages by building dendrograms based on the first transfer results. However none of the dendrograms produced were similar to the family hierarchy,which may indicate that either the differences we observe in our results match some other linguistic properties than the language families or that the differences are due to empirical differences in the data.

\section{Conclusion}
We successfully adapted Lim's \textit{et al.} semantic analogies approach \cite{lim2019}, applied it to morphological analogies, and achieved competitive results. Compared to the model of Lim \textit{et al.}, our CNN model is more flexible in many aspects:
    \textit{(i)} it is able to model any words even those never encountered in the training phase;
    \textit{(ii)} it is able to carry over domain and language specificities from the training process;
    \textit{(iii)} it has strong  potential to carry over models of analogy when using an adapted embedding model as shown with our transfer experiments;
    \textit{(iv)} it has strong potential to learn under different sets of axioms to adapt to other analogical settings. 
Our early experiments on transferability highlighted the potential to transfer and reuse our neural approach across domains. Therefore, we developed two bilingual models and a multilingual model that we applied across the 11 languages and tested different transferability settings.

%
%

\newpage
 \bibliographystyle{splncs04}
 \bibliography{mybibliography}

\newpage
\appendix

\section*{Appendix}


\begin{table}[ht]
    \centering
    \caption{Number of analogies for each language before data augmentation.}
    \vspace{5mm}
    \begin{tabular}{lrrr}\toprule
     \textbf{Language}   & \qquad \textbf{Train} & \qquad \textbf{Dev} & \qquad \textbf{Test} \\ \midrule
            Arabic              & 373240 & 7671 & 555312 \\
            Finnish             & 1342639 & 22837 & 4691453 \\
            Georgian            & 3553763 & 67457 & 8368323 \\
            German              & 994740 & 17222 & 1480256 \\
            Hungarian           & 3280891 & 70565 & 66195 \\
            Maltese             & 104883 & 3775 & 3707 \\
            Navajo              & 502637 & 33976 & 4843 \\
            Russian             & 1965533 & 32214 & 6421514 \\
            Spanish             & 1425838 & 25590 & 4794504 \\
            Turkish             & 606873 & 11518 & 11360 \\\bottomrule
    \end{tabular}
    \label{tab:sigmorphon}
\end{table}

\begin{figure}[ht]
\centering
\subcaptionbox{Hungarian - Finnish}[\textwidth]{\includegraphics[width=.65\textwidth]{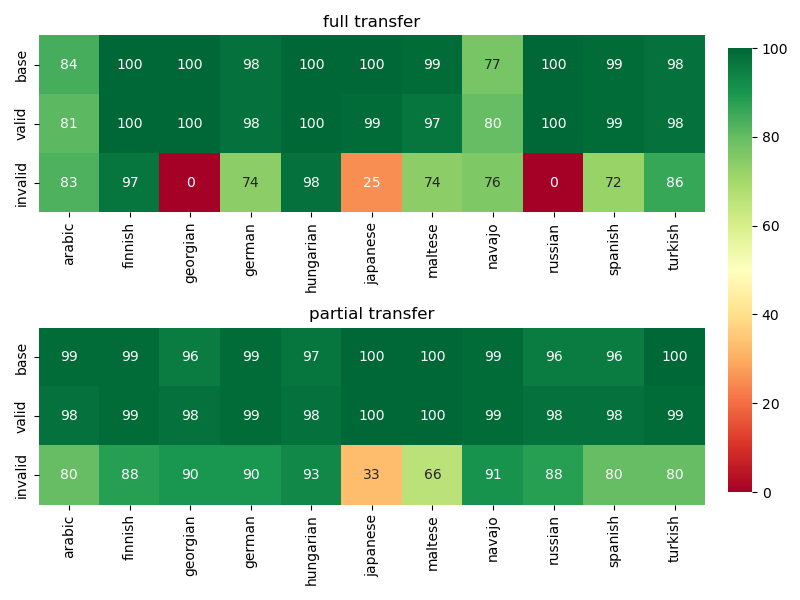}}
\subcaptionbox{Hungarian - Turkish}[\textwidth]{\includegraphics[width=.65\textwidth]{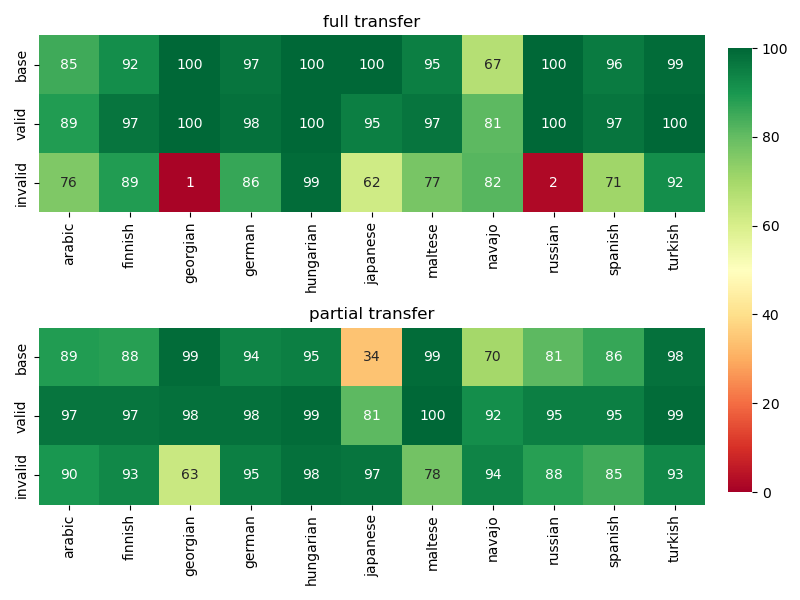}}
\caption{Accuracy (in \%) of fully transferred bilingual models on \sig and \japan.}
\label{fig:cmap-biling}
\end{figure}

\begin{figure}[ht]
\centering
\subcaptionbox{One embedding model}[\textwidth]{\includegraphics[width=.65\textwidth]{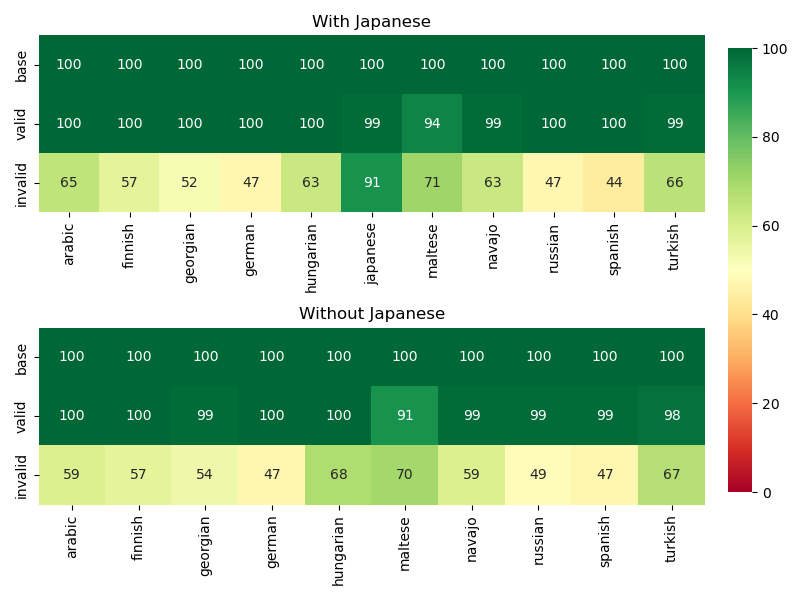}}
\subcaptionbox{Multiple embedding models}[\textwidth]{\includegraphics[width=.65\textwidth]{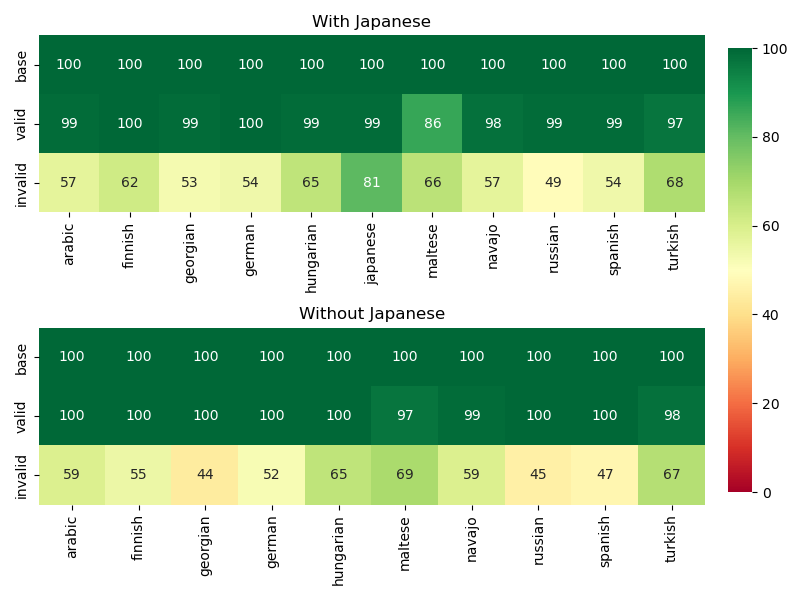}}
\caption{Accuracy (in \%) of fully transferred multilingual models on \sig and \japan.}
\label{fig:cmap-multiling}
\end{figure}

\begin{figure}[ht]
\centering
\subcaptionbox{Base}[\textwidth]{\includegraphics[width=.65\textwidth]{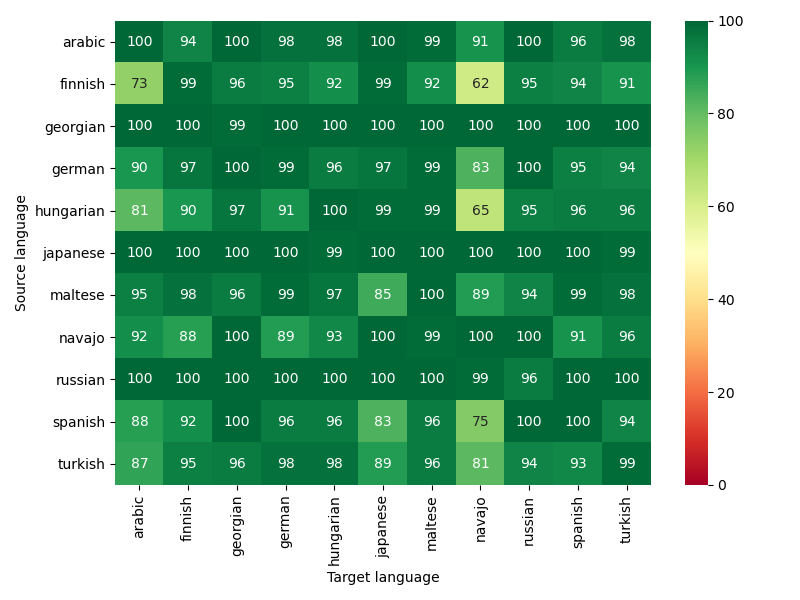}}
\subcaptionbox{Positive}[\textwidth]{\includegraphics[width=.65\textwidth]{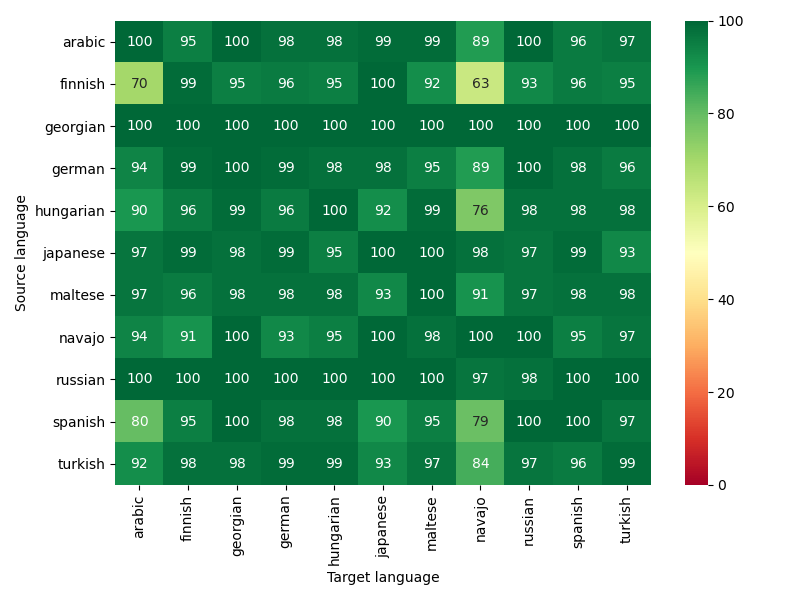}}
\subcaptionbox{Negative}[\textwidth]{\includegraphics[width=.65\textwidth]{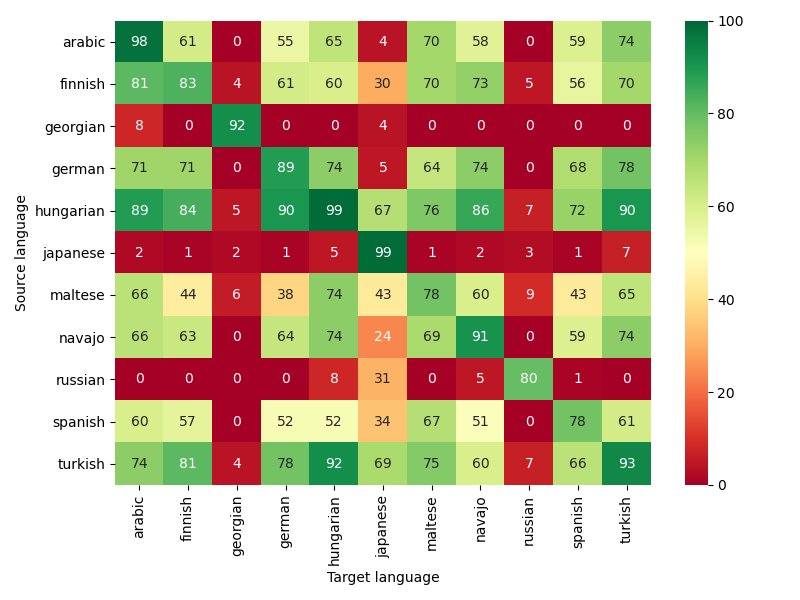}}
\caption{Accuracy (in \%) of fully transferred models on \sig and \japan.}
\label{fig:cmap-full}
\end{figure}

\begin{figure}[ht]
\centering
\subcaptionbox{Base}[\textwidth]{\includegraphics[width=.65\textwidth]{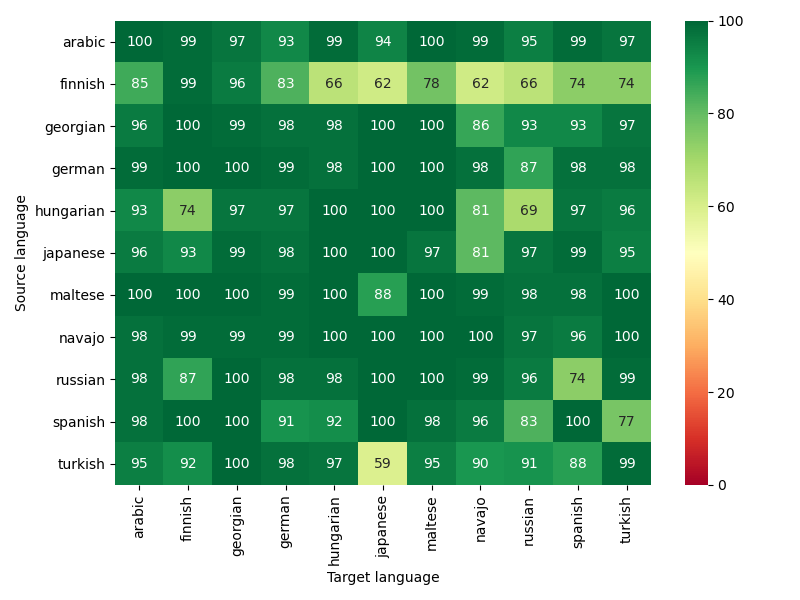}}
\subcaptionbox{Positive}[\textwidth]{\includegraphics[width=.65\textwidth]{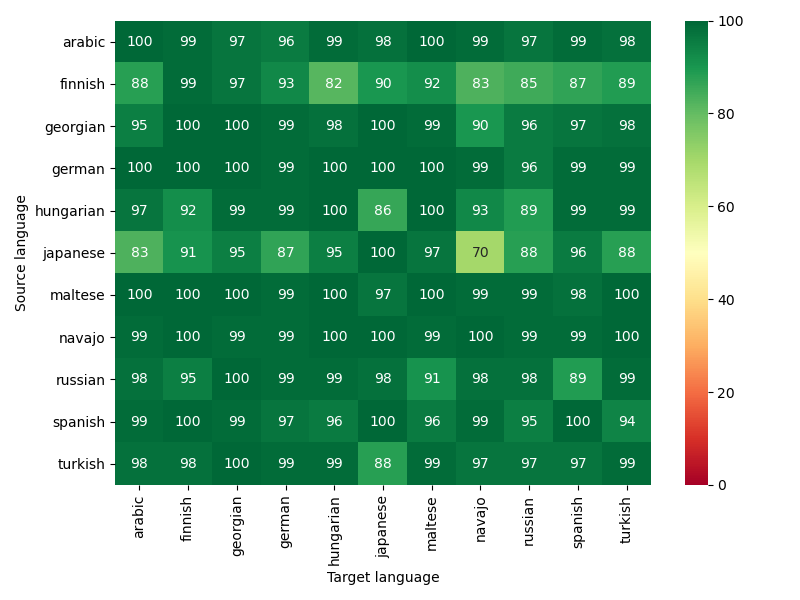}}
\subcaptionbox{Negative}[\textwidth]{\includegraphics[width=.65\textwidth]{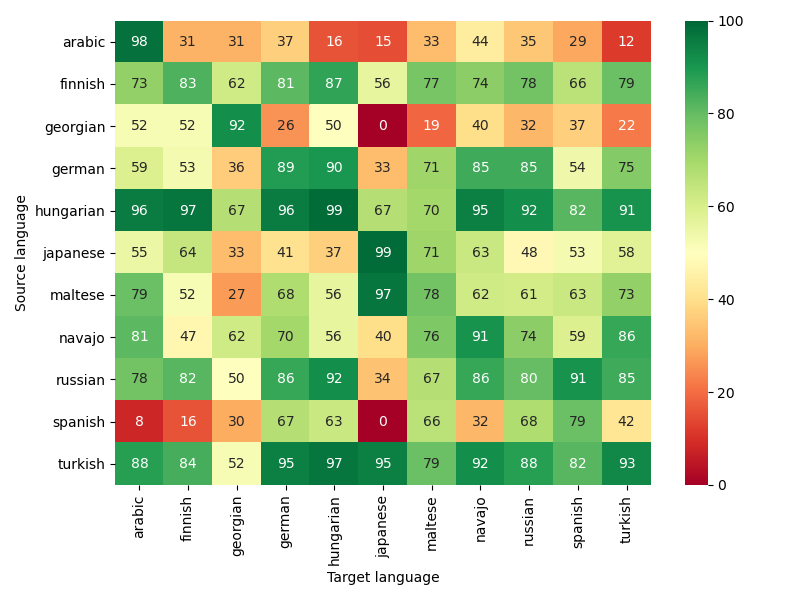}}
\caption{Accuracy (in \%) of partially transferred models on \sig and \japan.}
\label{fig:cmap-part}
\end{figure}

\end{document}